\pdfoutput=1
\documentclass[10pt,twocolumn,letterpaper]{article}

\usepackage[pagenumbers]{cvpr} 

\usepackage[table]{xcolor}
\usepackage{booktabs}
\definecolor{First}{RGB}{255,210,170}
\definecolor{Second}{RGB}{255,235,210}
\newcommand{\first}[1]{\cellcolor{First}\textbf{#1}} 

\newcommand{\second}[1]{\cellcolor{Second}#1}    

\newcommand{\legendfirst}{\colorbox{First}{\strut first}}
\newcommand{\legendsecond}{\colorbox{Second}{\strut second}}

\usepackage{booktabs}
\usepackage{multirow}
\usepackage[table]{xcolor}

\usepackage{graphicx}
\usepackage{subcaption}
\usepackage{adjustbox}
\usepackage{xcolor}

\definecolor{cvprblue}{rgb}{0.21,0.49,0.74}
\usepackage[pagebackref,breaklinks,colorlinks,allcolors=cvprblue]{hyperref}

\title{Data-Centric Visual Development for Self-Driving Labs}

\author{
Anbang Liu$^{\dagger,1}$ \quad
Guanzhong Hu$^{\S,2}$ \quad
Jiayi Wang$^{\dagger,3}$ \quad
Ping Guo$^{\S,4}$ \quad
Han Liu$^{\dagger,5}$\\[4pt]
$^{\dagger}$Department of Computer Science, Northwestern University\\
$^{\S}$Department of Mechanical Engineering, Northwestern University\\[4pt]
{\tt\small
$^{1}$anbangliu2027@u.northwestern.edu}\\
{\tt\small
$^{2}$guanzhonghu2028@u.northwestern.edu \quad
$^{3}$jiayiwang2020@u.northwestern.edu}\\
{\tt\small
$^{4}$ping.guo@northwestern.edu \quad
$^{5}$hanliu@northwestern.edu}
}

\begin{document}
\maketitle
\begin{abstract}
Self-driving laboratories offer a promising path toward reducing the labor-intensive, time-consuming, and often irreproducible workflows in the biological sciences. Yet their stringent precision requirements demand highly robust models whose training relies on large amounts of annotated data. However, this kind of data is difficult to obtain in routine practice, especially negative samples. In this work, we focus on pipetting, the most critical and precision sensitive action in SDLs. To overcome the scarcity of training data, we build a hybrid pipeline that fuses real and virtual data generation. The real track adopts a human-in-the-loop scheme that couples automated acquisition with selective human verification to maximize accuracy with minimal effort. The virtual track augments the real data using reference-conditioned, prompt-guided image generation, which is further screened and validated for reliability. Together, these two tracks yield a class-balanced dataset that enables robust bubble detection training. On a held-out real test set, a model trained entirely on automatically acquired real images reaches 99.6\% accuracy, and mixing real and generated data during training sustains 99.4\% accuracy while reducing collection and review load. Our approach offers a scalable and cost-effective strategy for supplying visual feedback data to SDL workflows and provides a practical solution to data scarcity in rare event detection and broader vision tasks.
\end{abstract}    
\section{Introduction}
\label{sec:intro}

\begin{figure*}[t]
  \centering
  \includegraphics[width=\linewidth]{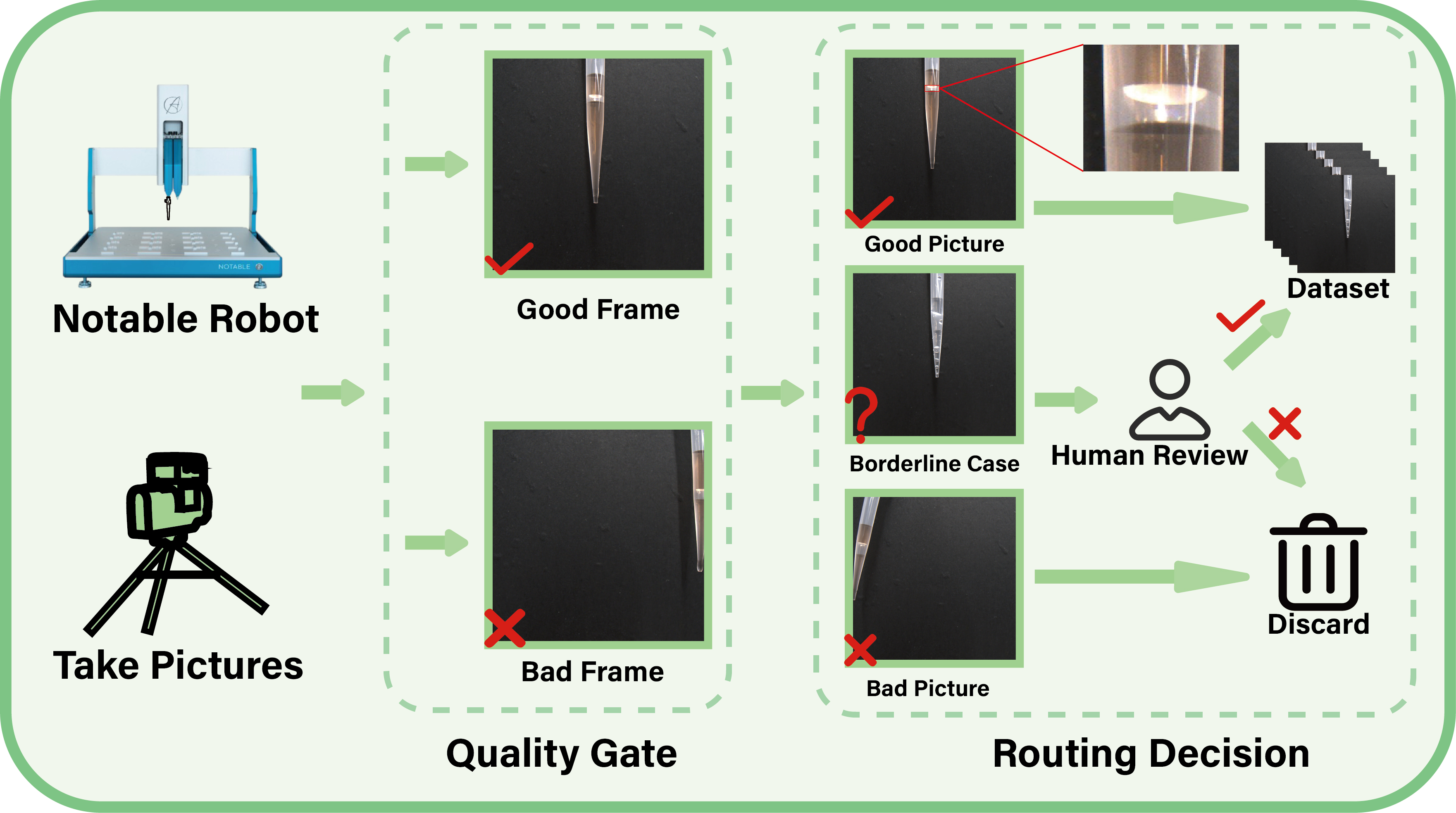}
  \caption{\textbf{Real track workflow.}
  After each aspiration, the robot holds the pipette tip at a fixed inspection place and the camera takes a photo. A quick quality check removes bad frames (e.g., off-center, or missing the tip). The remaining frames are screened by a lightweight classifier: good photos are accepted automatically, borderline cases are sent to a brief human review, and only poor-quality frames are discarded. Both bubble and no-bubble images are kept, so the process yields a steady, labeled stream of high-quality real data with minimal supervision.}
  \label{fig:real_track_flow}
\end{figure*}

Modern AI is powered by three engines—models, algorithms, and data—but in many practical settings the limiting factor is no longer architectural capacity. It is the availability of the right data, especially the negative or failure cases that determine reliability \cite{Cui19,Zhang21,Mumuni22,Wang24}. Self-driving laboratories (SDLs) illustrate this gap vividly \cite{Tobias25,Hysmith24}. Today’s SDL pipelines often lack visual feedback, so there is no closed-loop perception to catch errors during routine operations \cite{Tobias25,Hysmith24}. Adding such visual feedback is not a trivial matter of training “one more classifier”. It requires a repeatable way to produce sufficient, task-relevant images drawn from the same physical workflow the feedback will monitor \cite{Zepel20,Eppel20}. In pipetting—the backbone operation for most wet-lab protocols—one important failure situation is air bubbles inside pipette tips. These events are rare in competent operation, and images of them are scarce \cite{Yin24,Kim21,Hessenkemper22,Dunlap24}. As a result, performance in this setting is constrained by data rather than by the choice of model or algorithm \cite{Cui19,Zhang21}.

We cast the problem as visual development: building the data supply chain that makes visual feedback feasible in SDLs \cite{Mumuni22,Wang24}. The question is not “which detector is best,” but “how do we continuously produce the images a detector needs, with minimal human effort, and at a cost that scales?” We answer this by designing a \textbf{data-centric methodology} that turns pipetting itself into a steady source of training data for \textbf{binary classification of bubble presence} inside pipette tips. The methodology couples \textbf{two coordinated tracks (real and virtual)} so that scarcity in routine operation does not translate into scarcity at training time. On the virtual side we leverage modern generative paradigms and domain randomization to increase the prevalence of informative failures \cite{Goodfellow14,Ho20,Karras19,Radford21,Tobin17}. The design goals are straightforward: integrate perception into the existing physical workflow without disrupting throughput, concentrate human effort only where uncertainty requires it, and use generation strategically to raise the prevalence of error data.

Three obstacles motivate our design. First, rarity and imbalance: in well-run labs most aspirations are correct, so bubbles form a long tail that leads to inherently imbalanced datasets \cite{Cui19,Zhang21}. Second, subtlety and variability: bubbles occupy a small spatial extent, can be partly occluded by the meniscus, and present differently under changes in illumination, liquid color, tip geometry, and viewpoint \cite{Yin24,Kim21,Hessenkemper22}. Third, throughput: capturing and validating mistakes has historically required technician attention, which does not scale to the volumes that modern vision training regimes expect. Annotation hours—not GPU hours—set the ceiling \cite{Russakovsky14,Lin14}. Together, these factors create a data bottleneck.

Our methodology addresses these obstacles with two complementary tracks, both focused on data acquisition and selection. In the real track, as shown in Fig.~\ref{fig:real_track_flow}, we insert perception into the pipetting loop with a fixed camera and a programmable routine. After each aspiration event, the system performs event-triggered capture, applies lightweight prescreening with a simple classifier, and then routes by confidence: high-confidence cases are accepted automatically, while borderline cases are sent to human review. This concentrates expert time on uncertain examples and lets the pipeline operate around the clock with minimal supervision. Importantly, the task is formulated at the image level—\textbf{bubble presence}—so the pipeline does not rely on costly pixel-level annotation or mask drawing. Event-triggered capture and confidence-based routing convert sporadic snapshots from a liquid-handling robot into a continuous, quality-controlled stream of labeled real images.

\begin{figure*}[t]
  \centering
  \includegraphics[width=\linewidth]{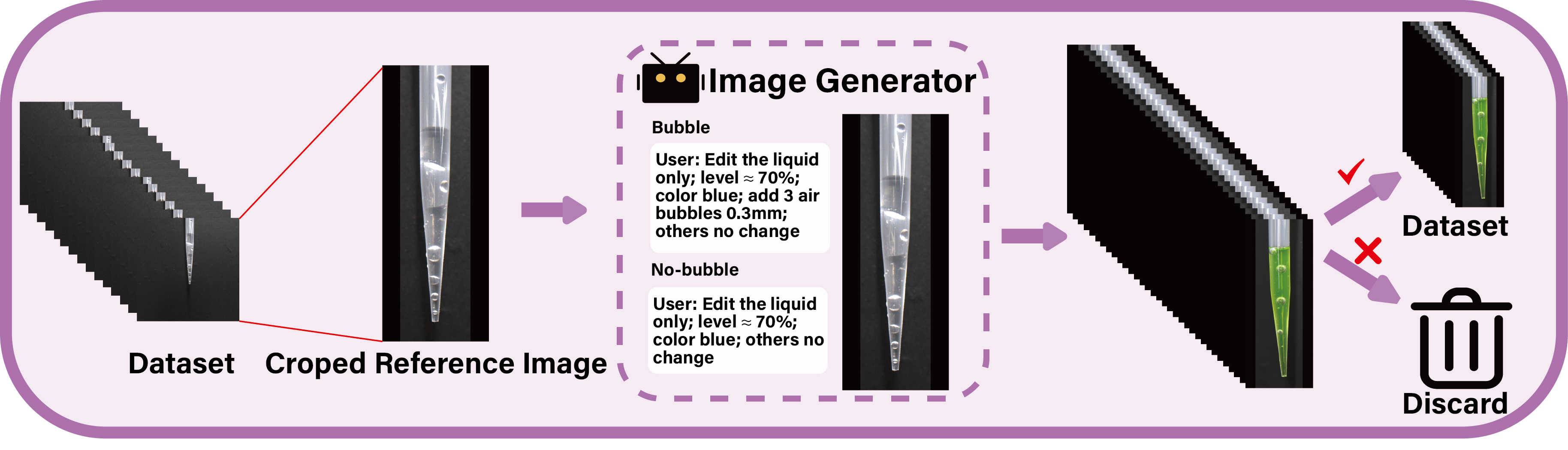}
  \caption{\textbf{Virtual track workflow.}
  Starting from a real reference tip image, we programmatically build prompts that fix viewpoint and background but vary lab factors (color, level, bubble count/size/distribution) and specify the intended class (bubble vs.\ no-bubble). We batch-generate variations, run the same quality gate as in the real track, enforce label consistency with the current classifier, and perform light human spot-checks. Both bubble and no-bubble images that pass are standardized to $600{\times}1500$ and added to the synthetic set for mixed training.}
  \label{fig:virtual_track_flow}
\end{figure*}

In the virtual track, as shown in Fig.~\ref{fig:virtual_track_flow}, we decide what to synthesize using priors from the physical setup. Real tip photos act as references, and prompts are derived from factors that matter in the lab—\textbf{liquid color, liquid level, bubble count, bubble size, bubble distribution, and lighting}. A modern text-guided image generator (Gemini 2.5 Flash Image) synthesizes images of liquid-containing tips with and without bubbles. Candidates are prescreened by the same classifier, and those that pass are finalized by human verification \cite{Goodfellow14,Ho20,Karras19,Radford21}. The objective is not to perfectly control bubble morphology—which remains stochastic—but to raise the prevalence of useful failure examples at very low marginal cost. Because prompts mirror physical variation and reference images anchor appearance, selected synthetic samples align with the downstream task and can be interleaved with real samples \cite{Tobin17}.

Both tracks feed a unified, class-balanced dataset used for training and evaluation. We provide standardized splits and training scripts so that others can reproduce our protocol and extend it to new checkpoints in the lab. For a concrete instantiation, we use an EfficientNetV2-L backbone for \textbf{binary classification of bubble presence} \cite{Tan21}. On a held-out real test set, a model trained only on automatically acquired real images attains 99.6\% accuracy while reducing technician time relative to manual collection. When we mix real and generated images during training, accuracy remains close to the real-only baseline (99.4\%) while further reducing real collection and manual effort \cite{Mumuni22,Wang24,Lin23}. The point is methodological: the feedback loop becomes viable not because the classifier is novel, but because the data engine supplies what the classifier needs, sustainably and at low cost.

Our study contributes to a broader data-centric perspective in AI for science and engineering. Instead of treating rare mistakes as insurmountable scarcity, we convert them into abundant supervision by coupling automation on the real side, and with \textbf{reference-conditioned, prompt-steered generation} on the virtual side \cite{Mumuni22,Wang24,Goodfellow14,Ho20}. The same recipe—automate reality and use physically guided generation to oversample failures—extends beyond bubble detection. It applies to other SDL visual checkpoints where errors are rare yet consequential, such as droplet misplacement, tip clogging, or cross-contamination traces \cite{Tobias25,Hysmith24,Zepel20}. It also applies to industrial visual inspection and other scientific imaging pipelines where staging faults is slow, expensive, or disruptive \cite{Zajec24,RamirezSanz23}. In these settings, the central artifact is not another network architecture, but a repeatable data process that turns operational pain points into scalable training signals.

Our contributions are fourfold:
\begin{enumerate}
    \item \textbf{Problem framing:} we formulate visual development for SDLs and identify why lack of visual feedback is a data supply problem rather than an architectural one, with bubble-in-tip classification as a concrete high-impact use case.
    
    \item \textbf{Real-world acquisition:} we design an \textbf{event-triggered, confidence-aware} collection loop that yields reliable real images with minimal manual intervention.
    
    \item \textbf{Physically guided generation:} we propose a \textbf{reference-conditioned, prompt-steered} synthesis strategy that selects synthetic images aligned with lab conditions and raises the prevalence of informative failures at low cost.
    
    \item \textbf{Unified dataset and evaluation:} we release standardized splits and training scripts and report results showing that real-only training achieves 99.6\% accuracy while mixed training maintains 99.4\% accuracy with reduced real collection and manual effort.
\end{enumerate}

Taken together, these elements offer a practical methodology for adding visual feedback to SDL workflows. By focusing on the \textbf{data engine}—how images are captured, selected, and combined—we enable scalable, low-cost data creation for rare-event visual quality control, moving practice toward the level of reliability that modern models promise but cannot reach without the right data \cite{Cui19,Zhang21}.

\section{Related Work}
\label{sec:relate}

\textbf{SDLs and Laboratory Automation.}
Self-driving laboratories (SDLs) aim to automate the scientific loop but most reports still emphasize planning, orchestration, and cloud execution rather than pervasive vision checkpoints inside unit operations \cite{Tobias25,Tom24,Hysmith24}. Vision-enabled stations exist for liquid-level control in chemistry setups \cite{Zepel20} and for task-specific automation platforms such as RoboCulture that integrate manipulation, sensing, and behavior trees for long-duration experiments \cite{Angers25}. In life-science automation, several works argue that liquid handlers lack integrated vision quality control, motivating computer-vision add-ons around accessible systems (e.g., OT-2) \cite{Khan25}. In robotics for aliquoting, YOLO-based perception has been used to guide manipulators \cite{Rybak23}, and AI models have been explored for liquid-level monitoring in assembly contexts \cite{Simeth21}. Compared to these lines, our focus is not a particular controller or station but the \emph{data supply chain} for a visual checkpoint (bubble/no-bubble) that SDLs currently miss, with an explicit mechanism to continuously create and curate training data from the pipetting loop itself.

\textbf{Vision for Transparent Containers, Liquids, and Bubbles.}
Datasets and methods for materials in transparent vessels (Vector-LabPics) demonstrate segmentation of vessels and phases but target general lab scenes rather than rare failure signatures inside pipette tips \cite{Eppel20}. LCDTC shifts toward liquid content estimation in containers using detection baselines \cite{Wu23}. Bubble research in two-phase or boiling flows explores segmentation and tracking (Mask R-CNN, SORT; BubbleID) \cite{Dunlap24}, robust detection under occlusion and overlap \cite{Hessenkemper22}, and generalized bubble mask extraction with weighted losses \cite{Kim21}. In pipetting contexts, recent work detects liquid retention in tips and proposes architectural tweaks to YOLOv8 for complex backgrounds \cite{Yin24}. We differ by (1) targeting the \emph{rare error} “air bubble in tip” as the supervision unit, (2) designing a bi-track engine to lift prevalence at data collecting time, and (3) releasing a balanced dataset where evaluation is on held-out \emph{real} tip images drawn from the same workflow.

\textbf{Data-Centric Learning and Imbalance.}
Long-tailed and imbalanced recognition motivates reweighting with effective numbers \cite{Cui19}, taxonomies and empirical syntheses of deep long-tailed learning \cite{Zhang21}, classic over/under-sampling (SMOTE-style) \cite{Chawla02}, and adaptive synthetic sampling (ADASYN) \cite{He08}. Streaming and drifting settings call for standardized evaluation across imbalance regimes \cite{Aguiar22}. Broad surveys argue for data-centric pipelines and augmentation beyond architectural changes \cite{Mumuni22,Wang24}. Our pipeline operationalizes these insights for SDL checkpoints: rather than solely reweighting a scarce minority, we \emph{manufacture} additional, task-aligned data through event-triggered collection and reference-conditioned synthesis.

\textbf{Synthetic Data and Generative Models.}
Generative modeling and representation learning (GANs, diffusion, vision–language) provide powerful tools to increase diversity \cite{Goodfellow14,Ho20,Karras19,Radford21}. Domain randomization shows that broad appearance variation can close sim-to-real gaps in robotic perception \cite{Tobin17}. For detection specifically, synthetic imagery can boost few-shot regimes, and CLIP can filter false positives from synthetic sets \cite{Lin23}. Our “virtual track” aligns with these trends but is intentionally \emph{reference-conditioned and prompt-steered}: real tip photos anchor appearance, prompts enumerate lab-relevant attributes (liquid color/level, bubble count/size/distribution, lighting), and a lightweight classifier plus human verification enforce task alignment before mixing with real data.

\textbf{Backbones and Detection Frameworks.}
Modern detectors and backbones form the toolset rather than the novelty in our work: EfficientNet/EfficientNetV2 for accuracy–efficiency scaling \cite{Tan19,Tan21}, residual networks \cite{He16}, ViT and hierarchical Swin Transformers \cite{Dosovitskiy20,Liu21}, and one-/two-stage detectors from YOLO \cite{Redmon16} to RetinaNet with focal loss for imbalance in dense detection \cite{Lin17}. Classical HOG and early deep features (DeCAF) contextualize the evolution of representations \cite{Dalal05,Donahue13}. We fix a single off-the-shelf classifier—EfficientNetV2-L—to isolate the data engine’s effect, and the \emph{bi-track pipeline} alone delivers strong real-set performance without bespoke architectures.

Prior work establishes why SDLs require vision and shows how liquids and bubbles can be detected in broader laboratory contexts. Research on long-tailed recognition and class imbalance explains why rare failures throttle reliability, and generative modeling offers a practical way to expand training data. We unify these threads into a practical, closed-loop \emph{data engine} for an SDL visual quality-control task—bubble-in-tip—by coupling automated real capture with prompt-steered, selection-based synthesis and confidence-guided human review.

\section{Method}
\label{sec:method}

\subsection{Overview}
We propose a bi-track data engine for bubble-in-tip perception in self-driving laboratory (SDL) workflows. In the \textbf{real track}, we integrate vision into pipetting by coordinating an ABLE Labs NOTABLE liquid-handling robot that performs pipetting, a fixed industrial camera that captures the tip immediately afterward, and a lightweight classifier that prescreens each image and routes ambiguous cases to human audit, enabling continuous 24/7 acquisition with minimal supervision. The \textbf{virtual track} complements scarcity by reference-conditioned, prompt-steered synthesis (Gemini 2.5 Flash Image), followed by classifier-consistency filtering and sparse human spot-checks. In both tracks we explicitly \emph{target both} bubble (\(y{=}1\)) and no-bubble (\(y{=}0\)) cases, filter \emph{unqualified} images (e.g., blur, framing, occlusion), not dispreferred classes. A standard EfficientNetV2-L classifier~\cite{Tan21} is trained with class-balanced loss to mitigate long-tail effects~\cite{Cui19}. This section formalizes three components: classifier and loss, real track, and virtual track.

\subsection{Classifier and Loss (EfficientNetV2-L)}
Let \(x\!\in\!\mathbb{R}^{H\times W\times 3}\) be an image; \(y\!\in\!\{0,1\}\) the label (1=bubble, 0=no-bubble); \(h_\theta(\cdot)\) the EfficientNetV2-L feature extractor~\cite{Tan21}; \((\mathbf w,b)\) the linear head; \(\sigma\) the sigmoid; \(\mathcal D\) the training set; \(n_y\) the sample count of class \(y\); \(\beta\!\in\![0,1)\) the class-balance hyperparameter.

\begin{equation}
  f_\theta(x) \;=\; \sigma\!\big(\mathbf w^\top h_\theta(x) + b\big).
\end{equation}
\begin{equation}
  \alpha_y \;=\; \frac{1-\beta}{1-\beta^{\,n_y}} \quad.
\end{equation}
\begin{equation}
\begin{aligned}
  \mathcal L_{\mathrm{CB}}
  \;=\; \frac{1}{|\mathcal D|}\!\!\sum_{(x,y)\in\mathcal D} \alpha_y\,
  &\Big[-y\log f_\theta(x)\\[-2pt]
  &-(1-y)\log\!\big(1-f_\theta(x)\big)\Big].
\end{aligned}
\end{equation}
After producing the posterior \(f_\theta(x)\) and computing the class weights \(\alpha_y\) via the effective-number formula, we minimize the weighted binary cross-entropy \(\mathcal L_{\mathrm{CB}}\) over \(\mathcal D\) to update \((\theta,\mathbf w,b)\).

\subsection{Real Track: Triggered Capture, Prescreen, Human-in-the-loop}
Let \(c_\theta(x)\) be the model confidence; \(\tau_A\) the auto-accept threshold; \(\tau_R\) the review threshold \((0.5\!\le\!\tau_R\!<\!\tau_A\!\le\!1)\); \(q(x)\!\in\![0,1]\) a scalar image-quality score (sharpness/framing heuristics) with threshold \(\tau_q\); \(\textsf{route}(\cdot)\) the router; \(h(\cdot)\) a human audit; \(\hat y(x)\) the assigned label.

\begin{equation}
  c_\theta(x) \;=\; \max\!\big\{\, f_\theta(x),\; 1-f_\theta(x) \,\big\}.
\end{equation}
\begin{equation}
\textsf{route}(x) \;=\;
\begin{cases}
  \text{A} & \text{if } q(x)\!\ge\!\tau_q \;\wedge\; c_\theta(x)\!\ge\!\tau_A,\\
  \text{R} & \text{if } q(x)\!\ge\!\tau_q \;\wedge\; \tau_R\!\le\! c_\theta(x)\!<\!\tau_A,\\
  \text{D} & \text{otherwise} \quad \text{(drop unqualified)}.
\end{cases}
\end{equation}
\begin{equation}
  \hat y(x) \;=\;
  \begin{cases}
    \mathbb{1}\!\big[f_\theta(x)\!\ge\!0.5\big], & \text{if } \textsf{route}(x)=\text{A},\\
    h(x), & \text{if } \textsf{route}(x)=\text{R}.
  \end{cases}
\end{equation}

After each capture, we first score image quality \(q(x)\) and model confidence \(c_\theta(x)\). The router then applies the thresholds: frames that satisfy \(q(x)\ge\tau_q\) and \(c_\theta(x)\ge\tau_A\) are auto-accepted and labeled by \(\mathbb{1}[f_\theta(x)\ge 0.5]\), frames with \(q(x)\ge\tau_q\) and \(\tau_R\le c_\theta(x)<\tau_A\) are sent to human audit \(h(x)\), and frames with \(q(x)<\tau_q\) are dropped. Accepted and audited samples form \(\mathcal D_{\mathrm{real}}=\{(x,\hat y(x))\}\). 

\begin{figure*}[t]
  \centering
  \captionsetup[subfigure]{justification=centering, font=small}

  \begin{subfigure}[t]{0.155\textwidth}
    \centering
    \includegraphics[width=\linewidth,height=2.5\linewidth,keepaspectratio]{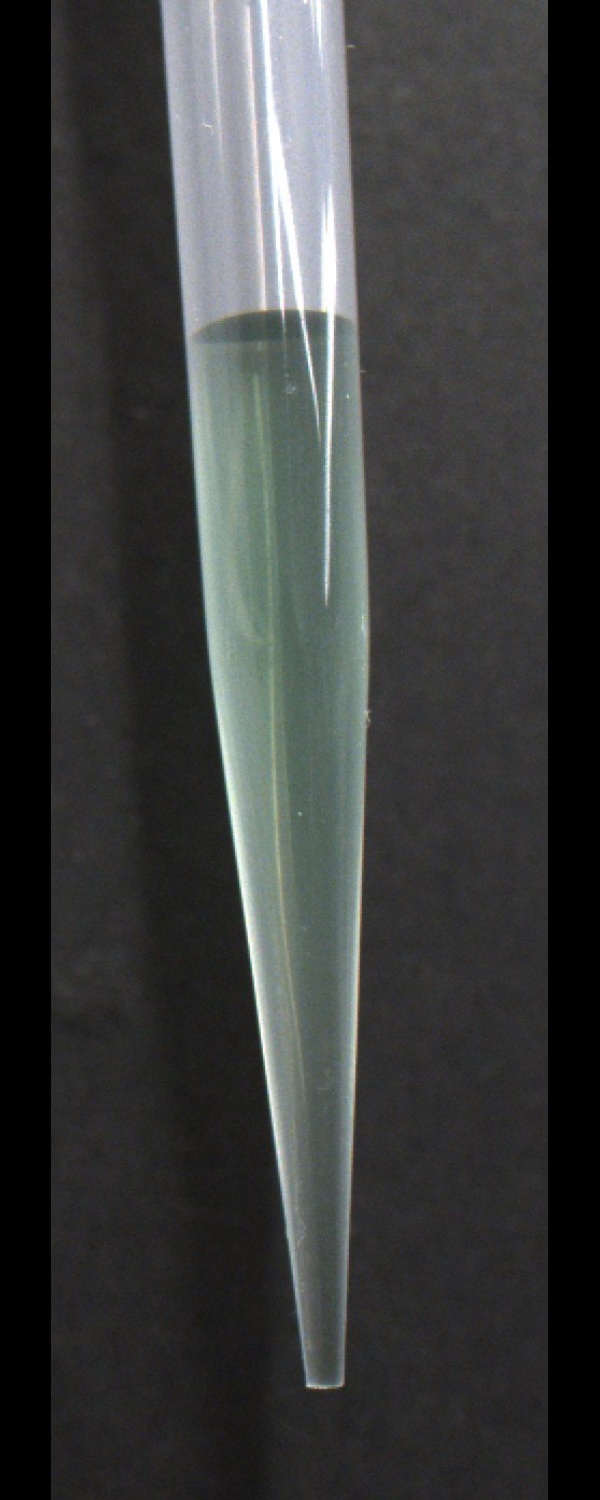}
    \caption{real \& no-bubble}
  \end{subfigure}\hfill
  \begin{subfigure}[t]{0.155\textwidth}
    \centering
    \includegraphics[width=\linewidth,height=2.5\linewidth,keepaspectratio]{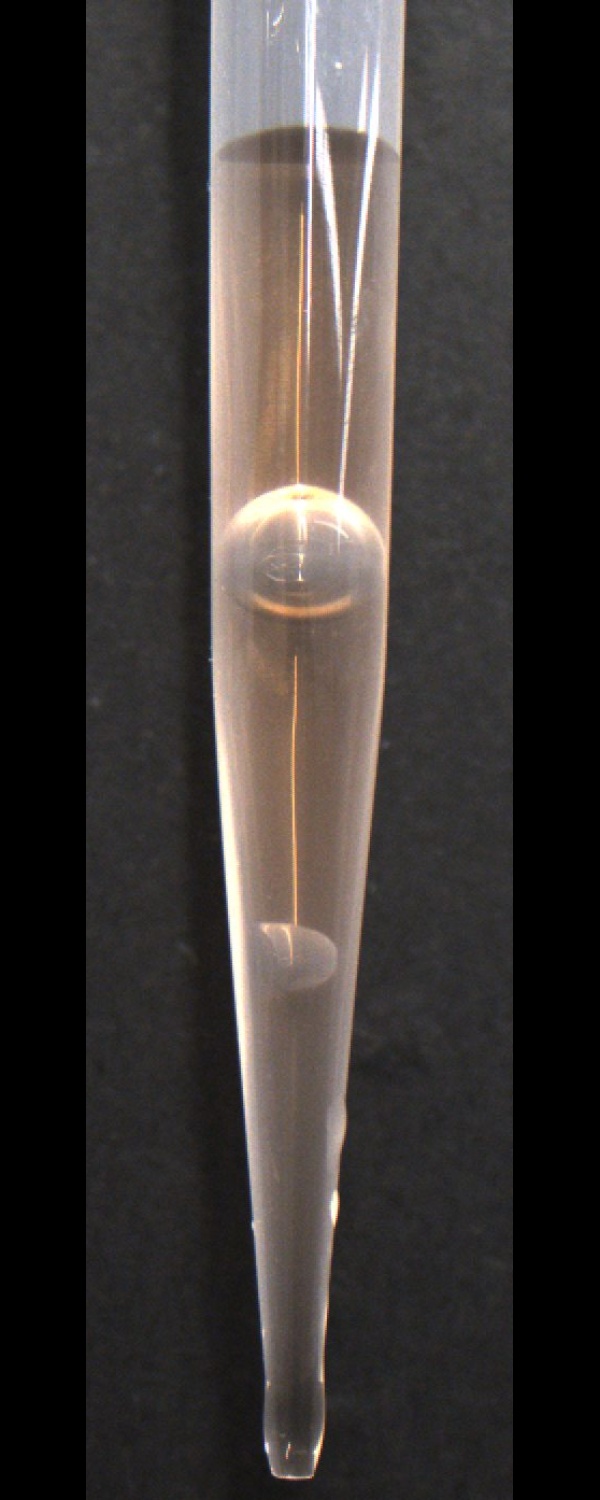}
    \caption{real \& bubble}
  \end{subfigure}\hfill
  \begin{subfigure}[t]{0.155\textwidth}
    \centering
    \includegraphics[width=\linewidth,height=2.5\linewidth,keepaspectratio]{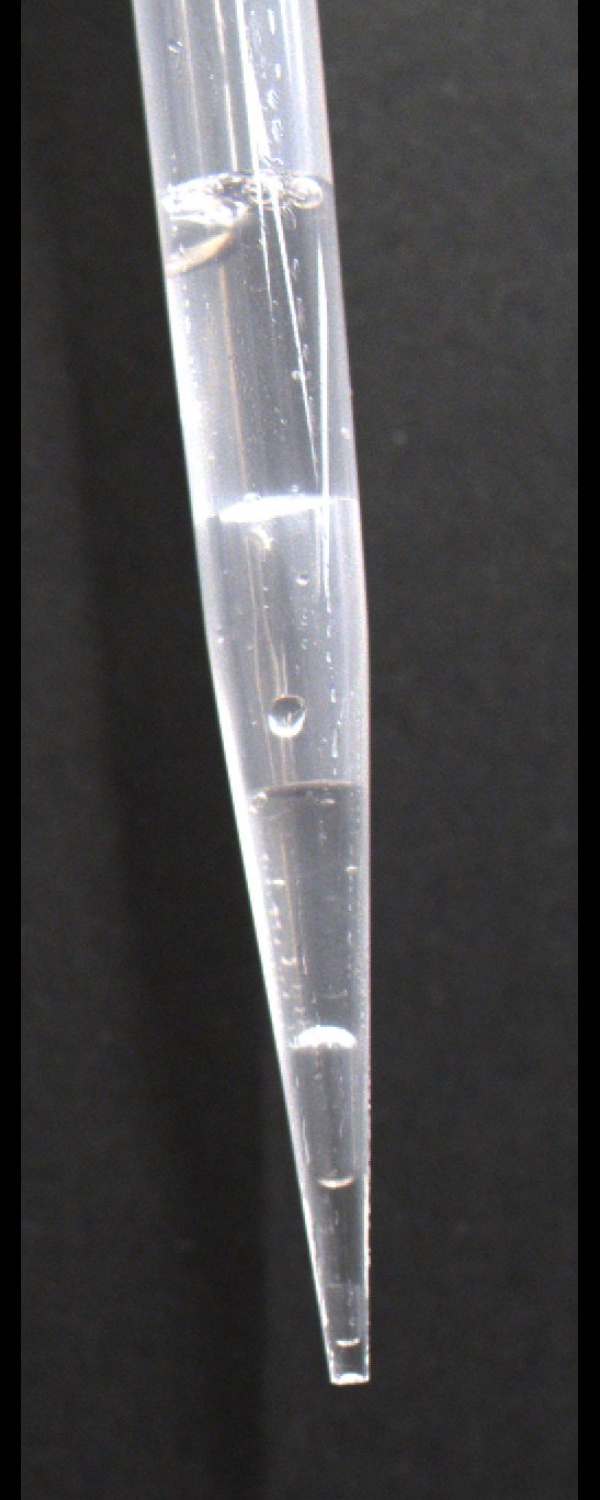}
    \caption{real \& bubble}
  \end{subfigure}\hfill
  \begin{subfigure}[t]{0.155\textwidth}
    \centering
    \includegraphics[width=\linewidth,height=2.5\linewidth,keepaspectratio]{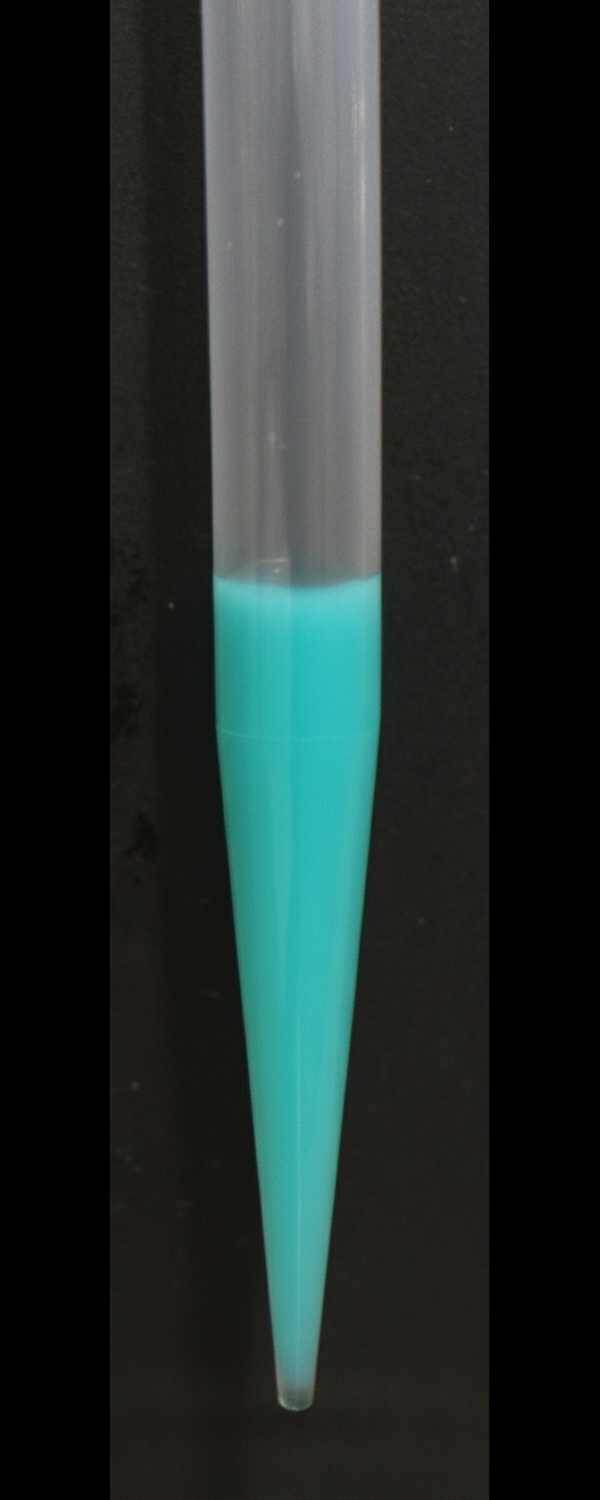}
    \caption{virtual \& no-bubble}
  \end{subfigure}\hfill
  \begin{subfigure}[t]{0.155\textwidth}
    \centering
    \includegraphics[width=\linewidth,height=2.5\linewidth,keepaspectratio]{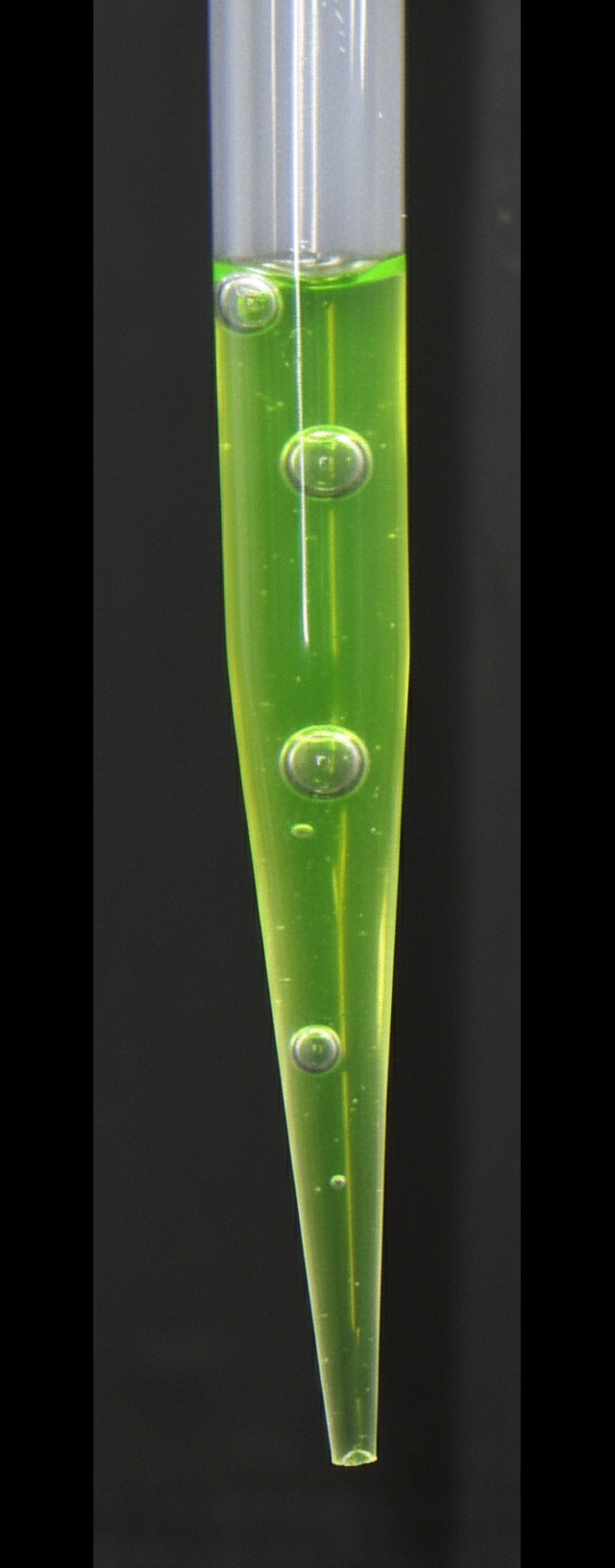}
    \caption{virtual \& bubble}
  \end{subfigure}\hfill
  \begin{subfigure}[t]{0.155\textwidth}
    \centering
    \includegraphics[width=\linewidth,height=2.5\linewidth,keepaspectratio]{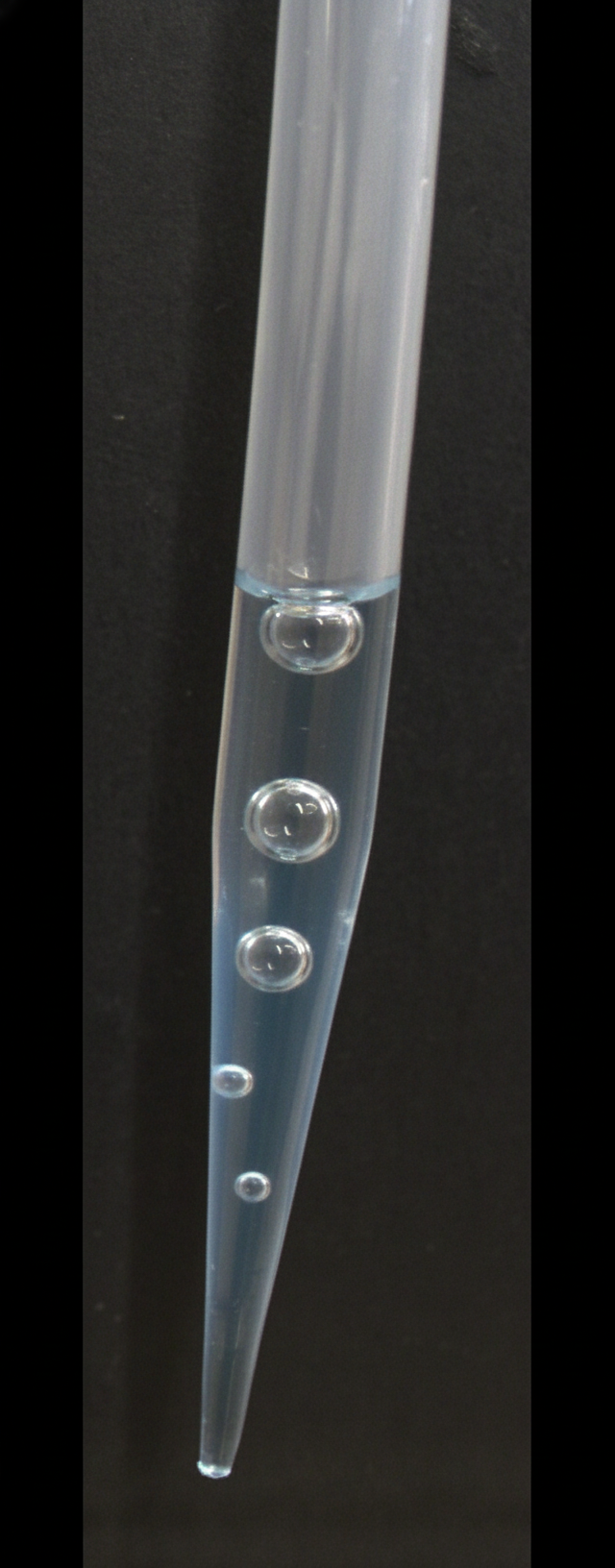}
    \caption{virtual \& bubble}
  \end{subfigure}

  \caption{\textbf{Dataset examples.}}
  \label{fig:dataset-strip}
\end{figure*}

\subsection{Virtual Track: Reference-Conditioned, Prompt-Steered Synthesis}
Let \(G_\psi\) be a text-guided image generator; \(z\!\sim\!p(z)\) noise; \(r\) a real tip reference image; \(\phi\) a prompt encoding lab factors (liquid color/level, bubble count/size/distribution, lighting) and the \emph{intended class}; \(\tilde x\) the synthesized image; \(\tilde y\!\in\!\{0,1\}\) the intended label from \(\phi\); \(\kappa_\theta\) a classifier-consistency score; \(\tau_k\) a keep threshold; \(\mathcal A_\rho\) a \(\rho\)-fraction spot-check set.

\begin{equation}
  \tilde x \;=\; G_\psi\!\big(z;\, r, \phi\big), 
  \qquad \tilde y \;=\; \ell(\phi)\in\{0,1\}.
\end{equation}
\begin{equation}
  \kappa_\theta(\tilde x,\tilde y) \;=\;
  \begin{cases}
    f_\theta(\tilde x), & \tilde y=1,\\
    1-f_\theta(\tilde x), & \tilde y=0.
  \end{cases}
\end{equation}
\begin{equation}
  \mathcal D_{\mathrm{syn}}
  \;=\;
  \{(\tilde x,\tilde y): \kappa_\theta\!\ge\!\tau_k \;\wedge\; q(\tilde x)\!\ge\!\tau_q \} \ \cup\ \mathcal A_\rho.
\end{equation}
Given a prompt \(\phi\) that specifies appearance factors and the \emph{intended} class \(\tilde y=\ell(\phi)\), we synthesize \(\tilde x=G_\psi(z;r,\phi)\), then compute the consistency score \(\kappa_\theta(\tilde x,\tilde y)\) and the quality score \(q(\tilde x)\). We keep a sample only if it passes both semantic consistency (\(\kappa_\theta\ge\tau_k\)) and basic quality (\(q\ge\tau_q\)), plus a small \(\rho\)-fraction spot-audit to control drift.

\subsection{Training on the Unified Set}
Let \(\delta,\epsilon\!\in\!(0,1]\) be sampling proportions for the real and synthetic pools, respectively. We form a proportioned union of (multi)sets as
\begin{equation}
  \mathcal D \;=\; \delta\,\mathcal D_{\mathrm{real}} \;\cup\; \epsilon\,\mathcal D_{\mathrm{syn}},
\end{equation}
where \(\delta\,\mathcal D\) denotes a subsample (or reweighted multiset) drawn from \(\mathcal D\) at proportion \(\delta\), targeting class balance across \(y\in\{0,1\}\). The training objective is
\begin{equation}
  \min_{\theta,\mathbf w,b}\ \mathcal L_{\mathrm{CB}}(\mathcal D).
\end{equation}

Results on held-out \emph{real} data validate that a mainstream backbone with a bi-track data engine attains very strong accuracy without bespoke architectures.

\section{Experiment}
\label{sec:exper}

\definecolor{BubbleCol}{HTML}{EAF6E8}    
\definecolor{NoBubbleCol}{HTML}{EAF1FF}  
\definecolor{LegendBG}{HTML}{F2F2F2}

\begin{table}[t]
\centering
\small
\setlength{\tabcolsep}{8pt}
\begin{tabular}{lccc}
\toprule
\multirow{2}{*}{\textbf{Source}} 
& \multicolumn{2}{c}{\textbf{Class}} 
& \multirow{2}{*}{\textbf{Total}} \\
\cmidrule(lr){2-3}
& \cellcolor{BubbleCol}\textbf{Bubble} 
& \cellcolor{NoBubbleCol}\textbf{No-bubble} 
& \\
\midrule
Real    
& \cellcolor{BubbleCol}1701 {\scriptsize(53.1\%)} 
& \cellcolor{NoBubbleCol}1501 {\scriptsize(46.9\%)} 
& 3202 \\
Virtual 
& \cellcolor{BubbleCol}1523 {\scriptsize(50.4\%)} 
& \cellcolor{NoBubbleCol}1499 {\scriptsize(49.6\%)} 
& 3022 \\
\midrule
\textbf{Overall} 
& \cellcolor{BubbleCol}\textbf{3224} {\scriptsize(52.0\%)} 
& \cellcolor{NoBubbleCol}\textbf{3000} {\scriptsize(48.0\%)} 
& \textbf{6224} \\
\bottomrule
\end{tabular}

\vspace{2pt}
\footnotesize
\begin{tabular}{@{}l@{}}
\colorbox{BubbleCol}{\strut\ \textcolor{black}{Bubble (1)}\ } \quad
\colorbox{NoBubbleCol}{\strut\ \textcolor{black}{No-bubble (0)}\ } \quad
\colorbox{LegendBG}{\strut\ \textcolor{black}{\% = within-source proportion}\ }
\end{tabular}

\caption{\textbf{Dataset composition.} Counts per source and class; percentages are computed within each source (row).}
\label{tab:data_counts}
\end{table}

\subsection{Datasets}
\label{sec:datasets}

\textbf{Scope and sources.}
As shown in \autoref{tab:data_counts}, our dataset contains \textbf{6{,}224} tip images comprising \textbf{3{,}202 real} captures and \textbf{3{,}022 virtual} renders. Real images are acquired by an ABLE Labs NOTABLE liquid-handling robot instrumented with a fixed FLIR camera. Each data is labeled for \emph{bubble presence} (binary). Virtual images are produced by reference-conditioned, prompt-steered synthesis (Gemini 2.5 Flash Image), with prompts specifying intended class (bubble/no-bubble) and appearance factors (liquid color/level, bubble count/size/distribution, lighting). All images are center-aligned crops at \textbf{600\,$\times$\,1500} px: when the source is taller than target we trim from the \emph{top} (preserving the meniscus region), and when wider we crop \emph{symmetrically} from left/right; if narrower, we letterbox pad to the target width. Dataset examples are shown in Fig.~\ref{fig:dataset-strip}.

\textbf{Composition.}
Real images cover two tip lengths (\emph{long}/\emph{short}) crossed with five colors (transparent, red, yellow, blue, green). Virtual images sample colors at random. Bubble count is uniformly specified from \textbf{1–15} and nominal bubble diameter is \textbf{$\sim$0.2–1.5\,mm} at capture scale. We keep both \emph{bubble} and \emph{no-bubble} classes in each track and filter out only \emph{unqualified} frames (e.g., blur, misframing) via an image-quality score before labeling.

\textbf{Splits and leakage control.}
We evaluate only on \emph{held-out real} images to avoid domain confounds. We perform \emph{random, stratified} splits by class: \textbf{Real} $\rightarrow$ \emph{train/val/test} $=$ \textbf{2242/480/480} images with per-class counts (Train: \emph{1191/1051}, Val: \emph{255/225}, Test: \emph{255/225} for bubble/no-bubble), and \textbf{Virtual} $\rightarrow$ \emph{train} $=$ \textbf{3022} images (Train: \emph{1523/1499}). The training \emph{pool} is the union of the real-train and synthetic-train pools, but the actual training set uses proportional subsampling from each source:
\[
\mathcal D_{\mathrm{train}} \;=\; S_{\alpha}\!\big(\mathcal D_{\mathrm{real}}^{\mathrm{train}}\big)\ \cup\ S_{\beta}\!\big(\mathcal D_{\mathrm{syn}}^{\mathrm{train}}\big),
\]
where $S_{\rho}(\cdot)$ selects a $\rho$-fraction without replacement (we report $\alpha,\beta$ with results). Validation and test sets are \emph{real-only} (real-val and real-test, respectively). We fix a global random seed for reproducibility.

\textbf{Labels and quality control.}
Ground truth for real images is binary—bubble present (1) or no-bubble (0). Frames that fail the quality gate (occluded tip, or off-center framing) are discarded. Ambiguous but otherwise qualified frames are routed to human audit before inclusion. Virtual images inherit an intended label from the prompt. Each candidate must pass a classifier-consistency check by the current model and the same quality gate, and we also perform sparse human spot-checks on a random subset. This policy ensures that the selection method removes only unqualified images rather than preferentially excluding either class.

\begin{figure*}[t]
  \centering
  \begin{subfigure}[t]{0.32\linewidth}
    \centering
    \includegraphics[width=\linewidth,height=5.2cm]{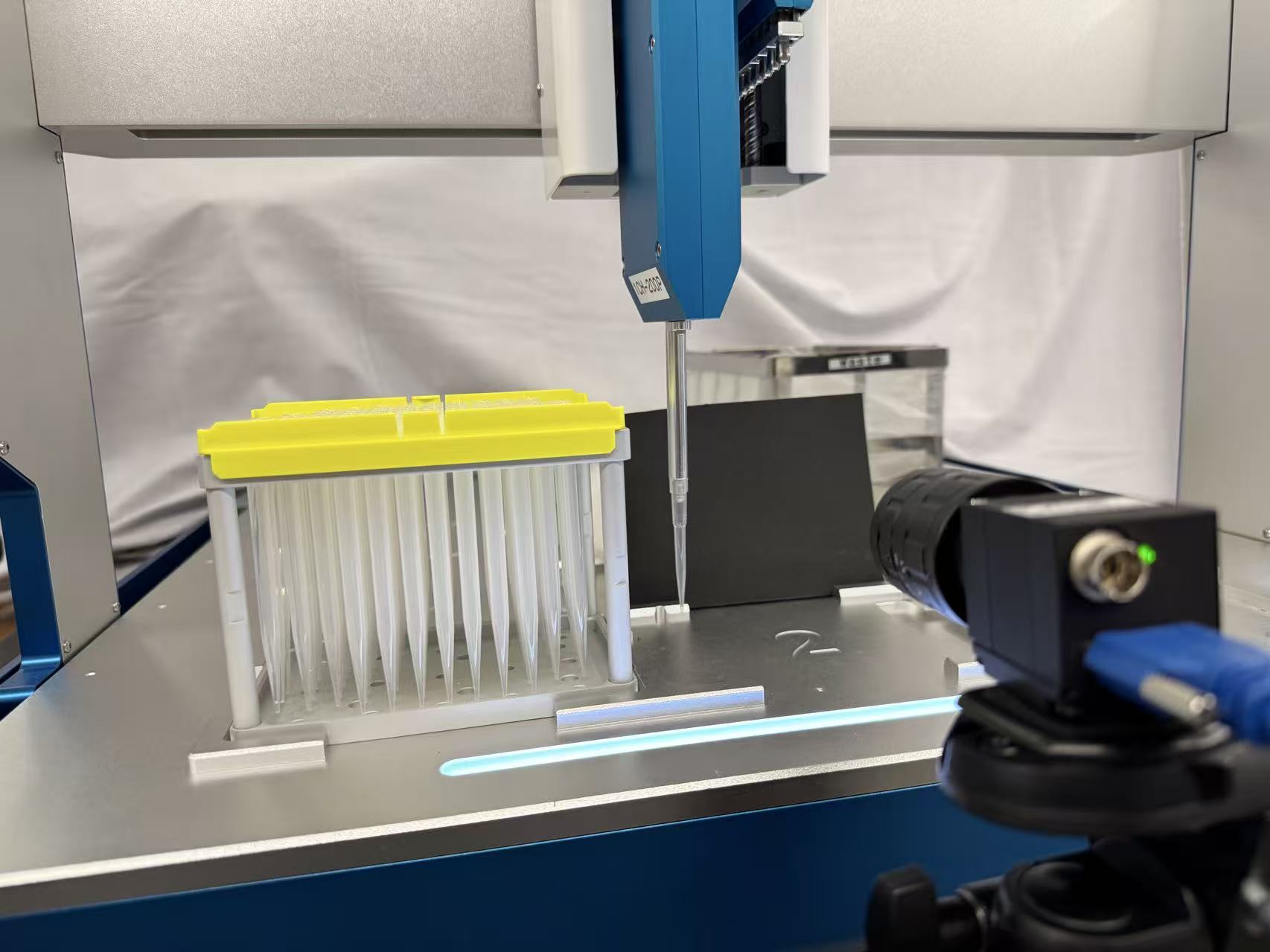}
    \caption{Left eye level}
  \end{subfigure}\hfill
  \begin{subfigure}[t]{0.32\linewidth}
    \centering
    \includegraphics[width=\linewidth,height=5.2cm]{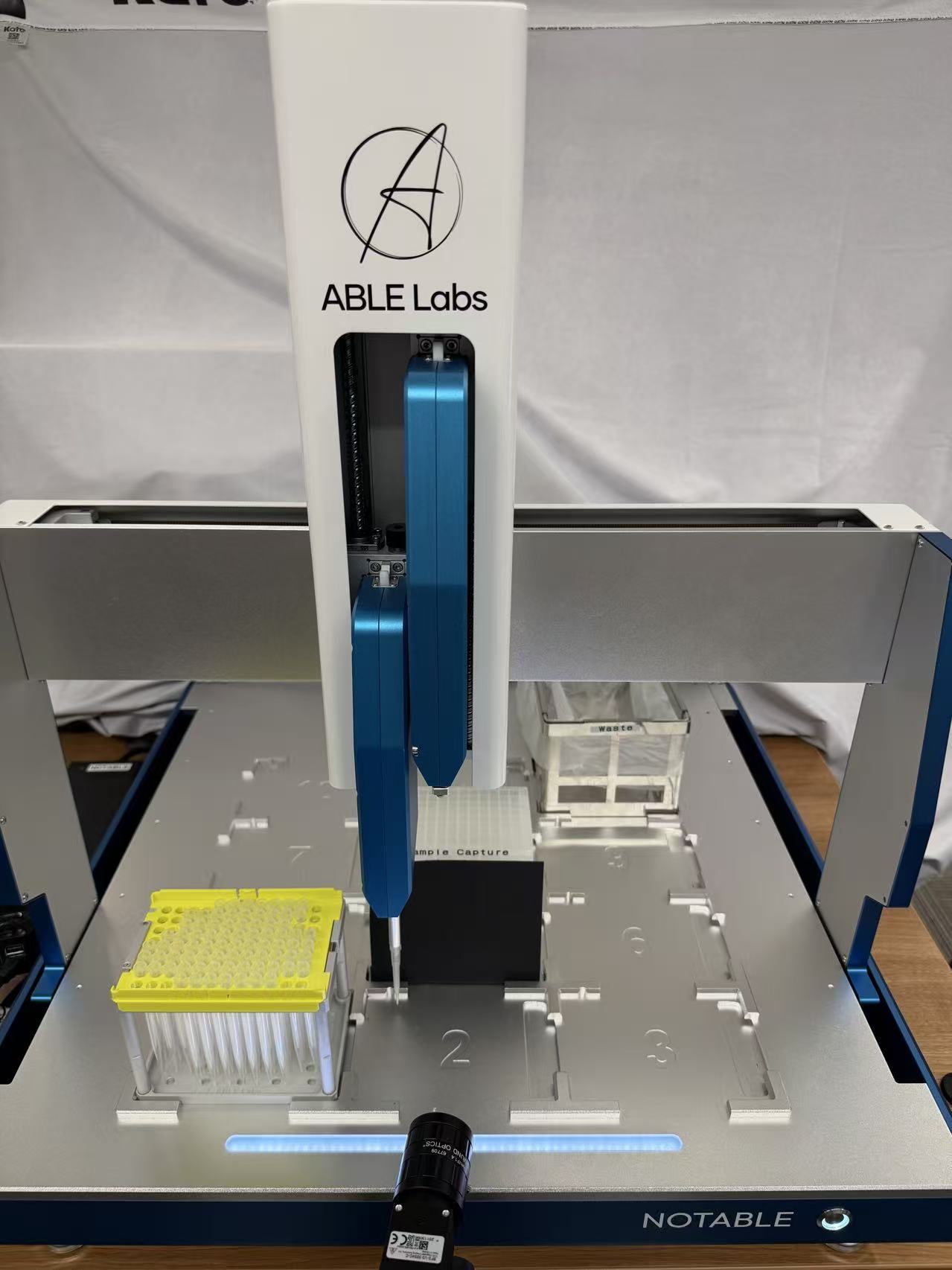}
    \caption{Top-down}
  \end{subfigure}\hfill
  \begin{subfigure}[t]{0.32\linewidth}
    \centering
    \includegraphics[width=\linewidth,height=5.2cm]{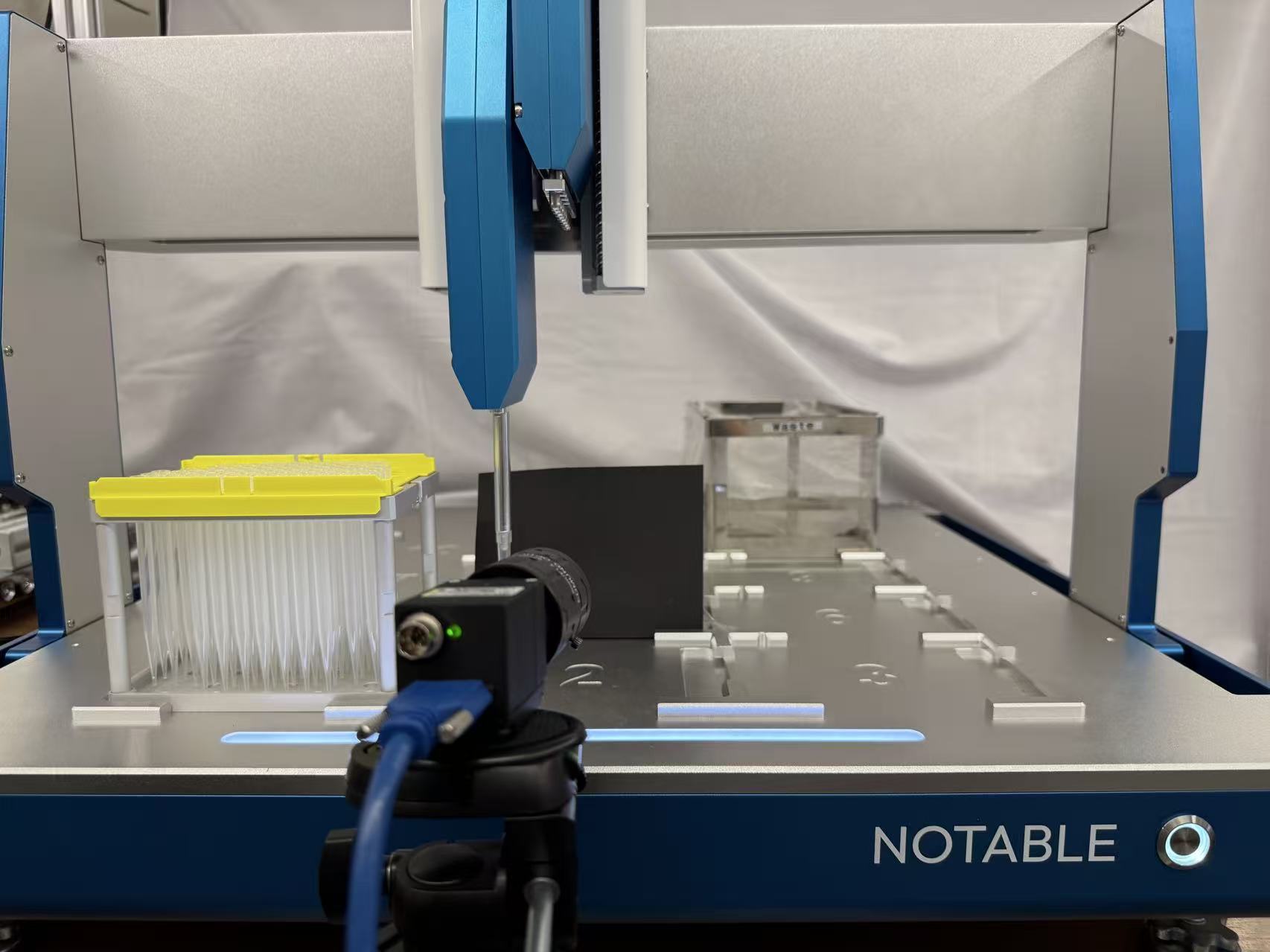}
    \caption{Right eye level}
  \end{subfigure}
  \caption{ABLE Labs \emph{Notable} liquid-handling robot and our fixed-camera setup from three viewpoints: (a) left eye level, (b) top-down, and (c) right eye level.}
  \label{fig:notable-views}
\end{figure*}

\subsection{Data Collection Setup}
\label{sec:setup}

\textbf{Hardware.}
As shown in Fig.~\ref{fig:notable-views}, we instrument an ABLE Labs \emph{Notable} liquid-handling robot with a fixed industrial camera--lens pair: a FLIR Blackfly S \emph{BFS-U3-50S4C-C} (USB3, color) and an Edmund Optics C-mount lens (6\,mm, F/1.4, \#67709). The camera is mounted on an external mount and rigidly aimed at the robot’s predefined inspection place (the repeatable capturing place after pipetting), so that every capture is taken from a nearly identical viewpoint. This out-of-workspace mounting avoids any interference with the robot’s motion. The chosen focal length and aperture provide a field of view that comfortably contains the full pipette tip and a small margin of surrounding background, while keeping the depth of field large enough that tips remain in focus despite minor height variations in the robot motion. We calibrate the mount orientation once and lock all adjustable joints, so that subsequent data collection sessions can be resumed without re-tuning the camera pose. In addition, we place a matte-black backdrop at the robot’s predefined inspection place to suppress background clutter and reflections, improving signal-to-noise in the tip ROI and contributing to the near-100\% accuracy model. This simple mechanical and optical setup favors robustness and reproducibility over complexity, and can be cheaply replicated in other laboratories without specialized vision hardware.

\textbf{Illumination.}
We use standard overhead LED lighting (neutral white, \(\sim\!4000\text{–}5000\,\mathrm{K}\)) with no strobes or auxiliary lights. This choice deliberately avoids carefully tuned studio lighting. Illumination settings, such as dimmer level or which ceiling fixtures are enabled, may be adjusted during a collection session to increase data diversity and expose the model to small changes in intensity and shadow patterns. We do not synchronize lighting with the robot or the camera, so the pipeline remains mechanically simple and does not depend on fragile trigger wiring or timing assumptions.

\textbf{Triggering \& timing.}
After the robot completes aspiration and holds the tip at a predefined inspection place, a Python program triggers the camera to acquire an image within \(<\!700\,\mathrm{ms}\) of motion stop to avoid motion blur. The trigger is issued by polling the robot state and firing only once the target location is satisfied, yielding a deterministic capture environment in the liquid-handling program. We use a fixed exposure time and gain configuration, so that the only major source of variability in the captured frames comes from the liquid interface and bubble patterns rather than from timing jitter. The end-to-end cycle time is dominated by liquid handling, and the vision stage adds \(<\!3\,\mathrm{s}\) per aspiration, which is small compared to other steps and therefore does not materially slow down the overall protocol.

\textbf{Quality gate.}
Frames are rejected immediately if they fail basic acquisition criteria: at first, we segment each captured image's tip region of interest (ROI) using Otsu’s thresholding. If no valid ROI is detected, the frame is discarded. Concretely, we apply a global threshold on the grayscale image, identify connected components corresponding to the bright tip against the black background, and keep only components that satisfy reasonable geometric constraints (such as minimum area and aspect ratio). Frames in which the tip is partially outside the field of view, too blurred to pass segmentation, or occluded by unexpected objects are thus filtered out before entering the manual or automated labeling pipeline. This lightweight “quality gate” can be run during collection and prevents corrupted observations from polluting the dataset or biasing downstream model evaluation.

\subsection{Processes}
\textbf{In the real track.}
We first calibrate the imaging geometry by fixing the FLIR camera and lens on an external mount, aligning the optical axis with the tip’s hold position, setting exposure and white balance. During collection, the ABLE Labs NOTABLE liquid handling robot executes aspiration and moves the pipette to a predefined inspection location, then a Python trigger acquires a frame within \mbox{$<\!700$\,ms} of motion stop. Each frame undergoes a quality gate that detects a valid tip ROI using OTSU-based segmentation, followed by geometric checks on ROI ratio, tip vertex angle, and bilateral side straightness via Hough lines. Frames that fail are dropped immediately. Qualified frames are then prescreened by a lightweight classifier and routed by confidence (auto-accept versus human review) to concentrate expert effort on borderline cases, yielding a reliable stream of both bubble and no-bubble exemplars without per-pixel annotation. This procedure embeds vision into wet-lab workflows and complements domain work on bubble and liquid perception in broader settings \cite{Eppel20,Kim21,Hessenkemper22,Dunlap24}.

\textbf{In the virtual track.}
To mitigate rarity, we synthesize reference-conditioned variations with Gemini~2.5 Flash Image in Batch (FILE) mode: (1) upload each real reference to the Files API to obtain a \texttt{file\_uri}; (2) build a JSONL where each line specifies a \textit{GenerateContentRequest} that fixes viewpoint/background while programmatically randomizing liquid color, fill level, and either bubble count in \mbox{$[1,15]$} or an explicitly bubble-free constraint; (3) submit a Batch job, poll to completion, then download and parse the results, saving returned images. Each candidate inherits an \emph{intended} label from the prompt (bubble/no-bubble) and is screened by the current classifier for label consistency and by the same quality gate as real data, with sparse human spot-checks. This selection-based synthesis leverages advances in generative modeling \cite{Goodfellow14,Ho20,Karras19,Radford21} and follows the spirit of domain randomization for robust transfer \cite{Tobin17}, while maintaining task alignment through reference conditioning. Related use of synthetic imagery for scarce-supervision detection is echoed in few-shot pipelines \cite{Lin23}.

\textbf{Data processing.}
All accepted images (real and synthetic) are standardized to \mbox{$600{\times}1500$}: if either side is smaller, we first upscale proportionally. If width exceeds 600\,px, we center-crop horizontally. If height exceeds 1500\,px, we top-crop to anchor the bottom of the tip, preserving the inspection region. We then apply light augmentations to improve robustness while respecting lab physics: small rotations (\(\pm 2^\circ\)), translation/crop jitter (\(\leq 3\%\)), brightness/contrast and gamma jitter to emulate LED variations, mild Gaussian noise/blur to model sensor and slight motion, and \emph{no} vertical flips (to avoid inverting gravity/meniscus). Class imbalance is handled at sampling and loss levels (e.g., effective-number weighting or minority oversampling) to keep bubble/no-bubble prevalence balanced per batch \cite{Cui19,Chawla02,He08}, and we train a standard EfficientNetV2-L classifier \cite{Tan21} on mixed mini-batches drawn from the curated real and synthetic pools according to the predefined mixing ratios.

\subsection{Results}

\begin{table}[t]
\centering
\scriptsize
\setlength{\tabcolsep}{4pt}
\begin{tabular*}{\linewidth}{@{\extracolsep{\fill}} lccccc}
\toprule
Mix (\%) & Train (Syn:Real) & Acc $\uparrow$ & Prec $\uparrow$ & Rec $\uparrow$ & F1 $\uparrow$ \\
\midrule
0   & 0:2240     & \first{0.9958} & \first{0.9961} & \first{0.9961} & \first{0.9961} \\
25  & 560:1680   & \first{0.9958} & \first{0.9961} & \first{0.9961} & \first{0.9961} \\
50  & 1120:1120  & \second{0.9938} & \second{0.9922} & \first{0.9961} & \second{0.9942} \\
75  & 1680:560   & 0.9917         & \second{0.9922} & \second{0.9922} & 0.9922 \\
\rowcolor{black!3}
100 & 2240:0     & 0.8503         & 0.8333         & 0.8984         & 0.8647 \\
\bottomrule
\end{tabular*}
\caption{Performance on a held-out \emph{real} test set under different synthetic:real training mixes (fixed budget 2{,}240). Higher is better ($\uparrow$). Top two per results are colored as \legendfirst\ and \legendsecond.}
\label{tab:mixing}
\end{table}

\textbf{Mixed training on a real test set.}
We ablate the proportion of synthetic images while keeping a fixed training budget of 2{,}240 images. Validation and test are \emph{real-only} and remain constant across runs. As shown in \autoref{tab:mixing}, accuracy is essentially unchanged when up to 25\% of the training set is synthetic. At 50–75\% synthetic, the drop is small (still $\geq$ 99.17\% accuracy). Training on 100\% synthetic images induces a marked domain gap. Precision dips slightly at 50\% synthetic, whereas recall remains high. This pattern suggests that selection-based synthesis preserves discriminative cues for bubbles but cannot fully substitute for real images. In practice, mixes in the 25–75\% range retain near-baseline accuracy while reducing the real collection load. This observation is consistent with prior findings that synthetic data works best as a complement rather than a replacement \cite{Lin23,Tobin17,Mumuni22,Wang24}.

\textbf{Cost, throughput, and acceptance.}
Our selection-based synthesis generates 3{,}600 candidates in about 30 minutes at a marginal cost of \$68. After classifier-consistency checks and the same quality gate used for real images, we retain 3{,}022 images (bubble and no-bubble combined), yielding an acceptance rate of \textbf{83.9\%}. This corresponds to \(\approx\)\textbf{\$0.0225} per accepted image and \(\sim\)\textbf{101} accepted images per minute. By contrast, the real track acquires roughly one frame every \(\sim\)10\,s, and no-bubble captures succeed nearly 100\% of the time, and deliberately producing bubble frames via careful pipetting succeeds only \(\sim\)\textbf{46\%}, which results in an effective \(\sim\)21\,s per \emph{accepted} bubble frame in addition to occasional human audits. The percentage of human audits is less than 10\%. Over the campaign we attempted $\sim$7,000 captures and retained 3,202 after quality gating and audit (1,701 bubble, 1,501 no-bubble), corresponding to an overall acceptance of $\sim$46\%. Synthesis therefore efficiently oversamples the rare failure class and stabilizes class balance at scale, consistent with reports that task-matched synthetic data, mixed judiciously, can deliver strong benefits with minimal downside \cite{Tobin17,Lin23,Mumuni22,Wang24}.


\section{Conclusion}
\label{sec:conclu}

We presented a data-centric methodology for visual development in self-driving labs that supplies the visual-feedback data lacking from pipetting workflows. The approach couples two coordinated tracks. The real track inserts perception into the physical loop with event-triggered capture, lightweight prescreening, confidence-based routing, and human review to yield reliable real images with minimal manual effort. The virtual track uses reference-conditioned, prompt-steered generation to synthesize liquid-filled tip images with and without bubbles, followed by prescreening and human verification, and both tracks feed a class-balanced dataset for training and evaluation. Experiments show that models trained on automatically acquired real data achieve 99.6\% accuracy on held-out real images, and that mixing real with generated data maintains 99.4\% accuracy while further reducing real collection and manual effort. Limitations include the stochastic control of synthetic appearance and evaluation on a single task. Future work will broaden visual quality-control tasks, improve controllability and calibration, and study domain shift. More broadly, the recipe offers a scalable, low-cost data supply chain for rare-event and general vision tasks.

\section{Code and Dataset Availability}
Code and dataset are available at 
\href{https://github.com/AndrewLiu666/Data-Centric-Visual-Development-for-Self-Driving-Labs}{\texttt{GitHub}}.
{
    \newpage
    \small
    \bibliographystyle{ieeenat_fullname}
    \bibliography{main}
}

\clearpage
\setcounter{page}{1}
\maketitlesupplementary

\section*{Supplementary: Prompt Templates for Virtual Data Generation}

\noindent
We generate images with Gemini~2.5 Flash Image in Batch (FILE) mode.
Below are the exact \emph{user} prompts sent to the model. Placeholders are
\verb|{liquid_color}|, \verb|{level_pct}|, \verb|{bubble_count}|.

\subsection*{Bubble-Present Prompt (parameterized)}
\noindent\textbf{User prompt:}
\begin{quote}\small\ttfamily
Use the provided reference photo of a pipette tip with bubbles to create a photorealistic variation. Only edit the liquid inside the tip. Set the liquid level to \(\sim\)\{level\_pct\}\% of the tip length. Set the liquid color to \{liquid\_color\} with realistic translucency/absorption matching the scene illumination. Insert \{bubble\_count\} small, realistic air bubbles inside the liquid column only; bubbles should be spherical to slightly oblate (\(\sim\)0.2--1.5 mm) with correct refraction, soft internal caustics, and specular highlights consistent with scene lighting. Distribute some near the inner wall/meniscus and some in the central volume. Keep the tip geometry, markings, background, camera viewpoint, exposure, depth-of-field, and sensor noise unchanged. Keep the meniscus physically plausible for the chosen level and color. Do not add foam, droplets on the exterior, text, or artifacts. Preserve the original image resolution (e.g., $600{\times}1500$) and cropping. Return only the final edited image; no text output.
\end{quote}

\subsection*{Bubble-Free Prompt (parameterized)}
\noindent\textbf{User prompt:}
\begin{quote}\small\ttfamily
Use the provided reference photo of a pipette tip to create a photorealistic variation. Only edit the liquid inside the tip. Set the liquid level to \(\sim\)\{level\_pct\}\% of the tip length. Set the liquid color to \{liquid\_color\} with realistic translucency/absorption matching the scene illumination. Remove all air bubbles inside the liquid column; the liquid must be perfectly bubble-free. Keep the tip geometry, markings, background, camera viewpoint, exposure, depth-of-field, and sensor noise unchanged. Keep the meniscus physically plausible for the chosen level and color. Do not add foam, droplets on the exterior, text, or artifacts. Preserve the original image resolution (e.g., $600{\times}1500$) and cropping. Return only the final edited image; no text output.
\end{quote}

\noindent
In both cases, we fix the camera viewpoint and background via the provided
reference image, randomize \verb|{liquid_color}| and \verb|{level_pct}|, and
for the bubble-present setting additionally randomize \verb|{bubble_count}|.

\end{document}